\documentclass[10pt,twocolumn,letterpaper]{article}
\pdfoutput=1
\usepackage[margin=1in]{geometry}
\usepackage[pagenumbers]{cvpr} 
\usepackage[table]{xcolor}
\definecolor{linkblue}{rgb}{0.21,0.49,0.74}
\usepackage[pagebackref,breaklinks,colorlinks,allcolors=linkblue]{hyperref}

\usepackage{booktabs,multirow,makecell}
\usepackage{array}
\usepackage{colortbl}
\usepackage{threeparttable}
\usepackage{algorithm}
\usepackage{algpseudocode}

\newcolumntype{d}{!{\vrule width 1.1pt}}

\newcommand{\best}[1]{\textbf{#1}}

\title{Generative Editing in the Joint Vision-Language Space for Zero-Shot Composed Image Retrieval}

\author{
    Xin Wang$^{1,2}$\thanks{Equal contribution.} \quad
    Haipeng Zhang$^{2}$\footnotemark[1] \quad
    Mang Li$^{2}$ \quad
    Zhaohui Xia$^{2}$ \\
    Yueguo Chen$^{1}$ \quad
    Yu Zhang$^{2}$\thanks{Corresponding authors.} \quad
    Chunyu Wei$^{1}$\footnotemark[2] \\[0.6em]
    {
        $^{1}$Renmin University of China \quad
        $^{2}$Alibaba Group
    }
}

\begin{document}
\maketitle
\begin{abstract}

Composed Image Retrieval (CIR) enables fine-grained visual search by combining a reference image with a textual modification. While supervised CIR methods achieve high accuracy, their reliance on costly triplet annotations motivates zero-shot solutions. The core challenge in zero-shot CIR (ZS-CIR) stems from a fundamental dilemma: existing text-centric or diffusion-based approaches struggle to effectively bridge the vision–language modality gap. To address this, we propose \textbf{Fusion-Diff}, a novel generative editing framework with high effectiveness and data efficiency designed for multimodal alignment. First, it introduces a \textbf{multimodal fusion feature editing} strategy within a joint vision-language (VL) space, substantially narrowing the modality gap. Second, to maximize data efficiency, the framework incorporates a lightweight \textbf{Control-Adapter}, enabling state-of-the-art performance through fine-tuning on only a \textbf{limited-scale} synthetic dataset of 200K samples. Extensive experiments on standard CIR benchmarks (CIRR, FashionIQ, and CIRCO) demonstrate that Fusion-Diff significantly outperforms prior zero-shot approaches. We further enhance the interpretability of our model by visualizing the fused multimodal representations. 
Together, our work establishes an effective, efficient, and interpretable generative paradigm for ZS-CIR tasks.

\end{abstract}


\section{Introduction}
\label{sec:intro}

Composed Image Retrieval (CIR) aims to retrieve a target image using a query composed of a reference image and a natural-language modification, enabling precise and user-controllable search beyond image-only or text-only retrieval~\cite{DBLP:conf/cvpr/BaldratiBUB22a, DBLP:conf/cvpr/ChenGB20, DBLP:conf/cvpr/HosseinzadehW20, DBLP:conf/cvpr/LeeKH21, DBLP:conf/iccv/0002OTG21, DBLP:conf/cvpr/Vo0S0L0H19}. Consider a fashion designer seeking ``this dress but in emerald green with long sleeves'' or an interior architect searching for ``this living room but with modern furniture.'' Such compositional queries capture how humans conceptualize visual search: building upon existing visual examples while articulating desired modifications. Despite its promise for open-world search, fashion recommendation, and e-commerce applications, CIR development has been constrained by two fundamental gaps: the \textit{data gap}—supervised methods rely on costly triplet annotations that are difficult to generalize~\cite{DBLP:conf/iclr/DelmasRCL22, DBLP:conf/aaai/0002YKK21}—and the \textit{modality gap}—aligning text-image compositions to visual targets remains challenging due to inherent semantic differences between textual and visual representations.

\begin{figure}[t]
    \centering
    \includegraphics[width=\linewidth]{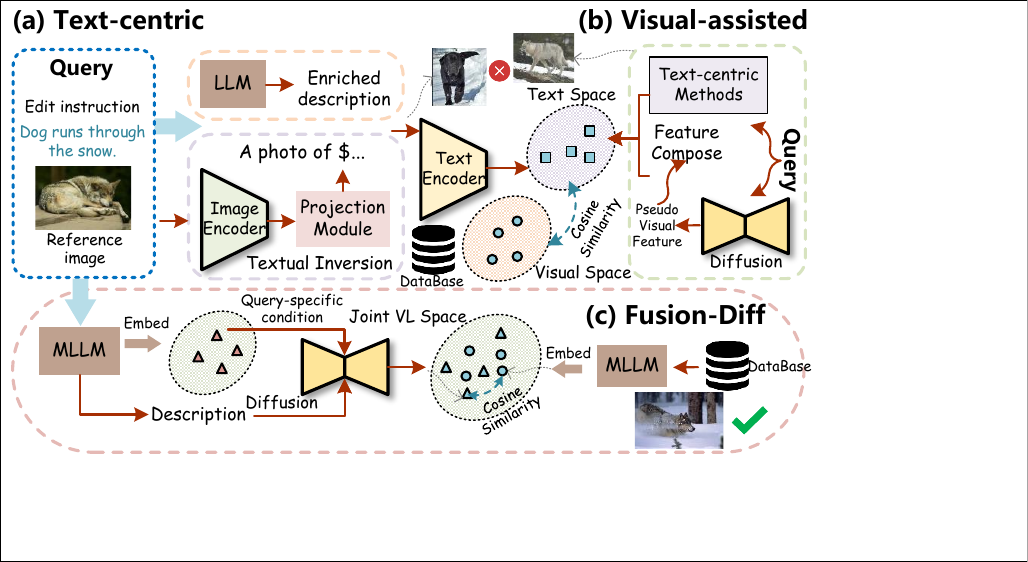}
    \caption{Comparison of different paradigms for zero-shot composed image retrieval. (a) \textbf{Text-centric methods} employ textual inversion or LLM-based description generation to perform retrieval in text space, discarding visual information. (b) \textbf{Visual-assisted methods} synthesize pseudo visual features via diffusion in visual space but still rely on text-centric retrieval mechanisms. (c) \textbf{Fusion-Diff} (Ours) operates directly in the joint vision-language space, modeling the distribution of target-fused embeddings to enable multimodal-to-multimodal retrieval, fundamentally addressing the modality gap.}
    \label{fig:introduction}
\end{figure}

Current Zero-Shot CIR (ZS-CIR) research has converged on several strategies to circumvent expensive triplet annotations. As illustrated in Figure~\ref{fig:introduction}(a), \textit{text-centric} approaches including \textit{inversion-based} methods~\cite{DBLP:conf/cvpr/SaitoSZLLSP23a, DBLP:conf/iccv/BaldratiA0B23, DBLP:conf/aaai/Tang0GZX0W24} map the reference image into the text space via textual inversion, simplifying the task to text-to-image retrieval but discarding fine-grained visual information. \textit{MLLM-based} methods~\cite{DBLP:conf/iclr/KarthikRMA24, DBLP:conf/cvpr/TangZQYG0LR0025, DBLP:conf/coling/BaoLX0025} leverage multimodal large language models (MLLM) to generate enriched textual descriptions, yet still perform cross-modal text-to-image matching. More recently, as shown in Figure~\ref{fig:introduction}(b), \textit{visual-assisted} generative approaches~\cite{DBLP:conf/cvpr/WangABL25, DBLP:conf/cvpr/LiMY25} synthesize pseudo-target visual features using diffusion models in the visual space, then compose them with text-centric methods for retrieval, but noise in synthesized features can misdirect retrieval focus.

Despite their differences, all existing methods share a fundamental limitation: they ultimately perform \textit{cross-modal} retrieval by matching text-based or text-enriched query representations against visual features, perpetuating the vision-language semantic gap. Text captures concepts through discrete symbolic representations, while visual features encode spatial, textural, and compositional information through continuous embeddings. This modality misalignment manifests as suboptimal retrieval performance, particularly for queries requiring fine-grained visual understanding. The core issue is the architectural choice to retrieve visual targets using query representations from a different modality space. This naturally inspires a pivotal question: \textit{could we instead perform retrieval within a unified joint vision-language space under the zero-shot setting?}


To address this challenge, we propose \textbf{Fusion-Diff}, a novel framework that performs retrieval by matching multimodal-to-multimodal representations rather than text-to-visual representations, as illustrated in Figure~\ref{fig:introduction}(c).

However, implementing this paradigm presents two critical challenges:
\begin{itemize}
\item \textbf{Distribution Modeling:} How can we effectively model the conditional distribution of fused multimodal representations without ground-truth target images during training? The multimodal fusion space exhibits intricate geometric structures where visual and textual semantics interact non-linearly. Moreover, the mapping from reference-modification pairs to target embeddings is inherently one-to-many: the same inputs can correspond to multiple valid interpretations depending on compositional ambiguity. Standard diffusion conditioning mechanisms prove insufficient, struggling to disentangle modification semantics from reference structure and model the uncertainty inherent in compositional interpretation.

\item \textbf{Data Efficiency:} How can we train such a diffusion model without massive-scale annotated triplet datasets, which would undermine the zero-shot advantage? Existing diffusion-based methods like CompoDiff~\cite{DBLP:journals/tmlr/GuCKJKY24} rely on billions of synthetic triplets, incurring substantial computational costs. The challenge lies in achieving strong feature editing capabilities while training on limited data, performing meaningful edits in high-dimensional multimodal feature space without overfitting or losing generalization across diverse domains.
\end{itemize}

To address the distribution modeling challenge, we introduce a conditional diffusion prior that operates directly in the joint vision-language space learned by pre-trained MLLMs. Our framework leverages an MLLM to generate target-oriented textual descriptions while producing fused embeddings from reference images and modification texts. We then learn a conditional distribution over target-fused embeddings, enabling us to sample multiple target-aware hypotheses at inference and perform ensemble retrieval in the multimodal pathway, effectively mitigating the vision-language gap while maintaining semantic consistency with both reference structure and modification instructions.

To address the data efficiency challenge, we introduce a lightweight \textbf{Control-Adapter} architecture that enables effective feature editing with minimal training data. Instead of training from scratch on billions of samples, our Control-Adapter is a parameter-efficient module that adapts a pre-trained diffusion backbone to the compositional editing task. We construct a compact synthetic dataset of only 200K triplets by mining semantically related image pairs from existing datasets and generating attribute-focused modification texts. The Control-Adapter injects compositional conditioning at multiple diffusion stages while preserving the generalization capability of the pre-trained backbone, allowing Fusion-Diff to achieve state-of-the-art performance with two orders of magnitude less training data than prior diffusion-based methods.

Our contributions are: (1) We propose the first framework to perform zero-shot composed image retrieval by modeling and sampling from the distribution of target-fused embeddings in a joint vision-language space, fundamentally addressing the modality gap. (2) We introduce a conditional diffusion prior over multimodal-fused features that captures the complex distribution of potential targets. (3) We design a lightweight Control-Adapter achieving strong feature editing with only 200K synthetic samples, demonstrating superior data efficiency. (4) Extensive experiments show that Fusion-Diff substantially outperforms existing zero-shot baselines while providing interpretable multimodal features.
\section{Related Work}
\label{sec:related_work}
\subsection{Zero-Shot Composed Image Retrieval}
Composed Image Retrieval facilitates fine-grained image search by combining a reference image with a textual edit. While early methods relied on fully supervised training with $(I_r, T_{\Delta}, I_t)$ triplets, the high cost of annotating such datasets has motivated a shift towards zero-shot CIR, which does not require triplet supervision. One prominent direction is inversion-based methods~\cite{DBLP:conf/cvpr/SaitoSZLLSP23a}~\cite{DBLP:conf/iccv/BaldratiA0B23}. These approaches function by projecting the reference image into the text space, creating "pseudo-tokens" that are then combined with the edit description. This compression of visual information into a textual format, however, struggles to retain fine-grained details. A contrasting strategy leverages Multimodal Large Language Models (MLLMs)~\cite{DBLP:conf/iclr/KarthikRMA24}~\cite{DBLP:conf/cvpr/TangZQYG0LR0025}~\cite{DBLP:conf/coling/BaoLX0025}. Here, the MLLM reasons over the composed query to generate a completely new, rich textual description of the target, which then directs the search. Despite their architectural differences, a common limitation persists: both paradigms enrich the query primarily on the textual side, failing to incorporate explicit visual evidence of the target and thus leaving the core modality gap unaddressed.

\subsection{Multimodal Fusion and Joint Vision-Language Spaces}
Vision-Language (VL) representation learning has evolved significantly. Initial works, exemplified by CLIP~\cite{DBLP:conf/icml/RadfordKHRGASAM21} and ALIGN~\cite{DBLP:conf/icml/JiaYXCPLLD21}, utilized a dual-encoder, late-fusion paradigm, which cannot represent compositional concepts in a single vector. Subsequent MLLMs, from BLIP-2~\cite{DBLP:conf/icml/0008LSH23} to PaLI-3~\cite{DBLP:journals/corr/abs-2310-09199}, employed deep cross-modal fusion, but their representations are optimized for \textit{text generation}, not as explicit, retrievable embeddings. A more recent consensus, seen in works like ImageBind~\cite{DBLP:conf/cvpr/GirdharELSAJM23}, UNIVERSAL~\cite{DBLP:journals/corr/abs-2402-09415}, MM-EMBED~\cite{DBLP:journals/corr/abs-2403-09551}, and GME~\cite{DBLP:conf/cvpr/ZhangZXLDLXZLZ25}, advocates for a unified joint embedding space that holistically models the static joint distribution. While these advances validate the unified space, they do not address the core CIR challenge: modeling the \textit{semantic transformation} from a query to a target. Our Fusion-Diff is the first to propose a conditional diffusion model operating directly within this unified VL space. This allows us to perform retrieval via multimodal-to-multimodal matching, directly addressing the modality gap that plagues prior ZS-CIR methods.

\subsection{Diffusion Models and Controllable Generation}
Diffusion Models have become state-of-the-art in generative tasks~\cite{DBLP:conf/nips/HoJA20, DBLP:conf/iclr/SongME21, DBLP:conf/iclr/ChenYGYXWK0LL24, DBLP:journals/corr/abs-2501-18427}, producing high-fidelity images conditioned on various inputs. Latent diffusion models~\cite{DBLP:conf/cvpr/RombachBLEO22} made this process more efficient by operating in a compressed latent space, driving tasks from T2I synthesis~\cite{DBLP:conf/nips/SahariaCSLWDGLA22, DBLP:conf/iccv/ZhangRA23} to editing~\cite{DBLP:conf/cvpr/BrooksHE23, DBLP:conf/iclr/MengACE22} and personalization~\cite{DBLP:conf/nips/SahariaCSLWDGLA22}. This generative-feedback paradigm was adopted in CIR~\cite{DBLP:journals/tmlr/GuCKJKY24, DBLP:conf/cvpr/WangABL25, DBLP:conf/cvpr/LiMY25}, where pseudo-target visual information is synthesized. This two-stage approach, however, is computationally expensive and critically bottlenecks retrieval performance on generative artifacts. Our Fusion-Diff adapts diffusion for a novel task: generative modeling in joint vision-language space. We are the first in ZS-CIR to learn a conditional diffusion prior directly on the distribution of the unified joint VL space. By sampling feature hypotheses rather than pixels, we bypass pixel-space pitfalls and enable robust, multimodal matching.
\section{Methodology}
\begin{figure*}
    \centering
    \includegraphics[width=1\linewidth]{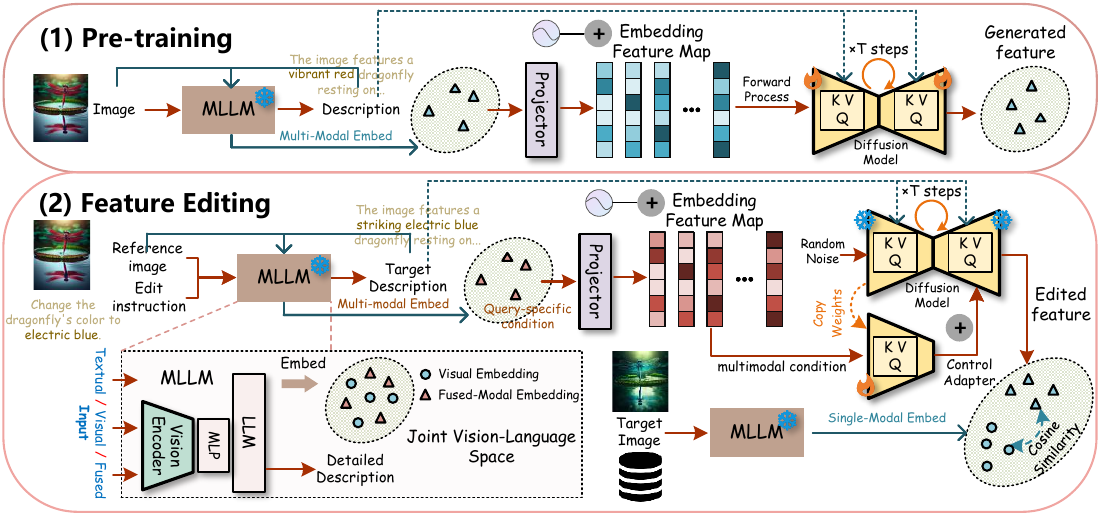}
    \vspace{-1ex}
    \caption{The Framework of Fusion-Diff.}
    \label{framework}
    \vspace{-1ex}
\end{figure*}

To resolve the modality gap identified in Sec.~\ref{sec:intro}, we propose Fusion-Diff, which performs retrieval within a joint vision-language space. Our method overcomes the key challenges of distribution modeling and data efficiency via a bifurcated training strategy: (1) we pre-train a conditional diffusion prior to model the distribution of joint vision-language space, and (2) we fine-tune with a lightweight, data-efficient Control-Adapter for the compositional edit task. An overview of our architecture is depicted in Figure ~\ref{framework}.

\subsection{Joint Vision-Language Space Preliminaries}
We adopt a multimodal large language model denoted by $f_M$ that can accept images, text, or image-text pairs as input. For any input $x$, we first construct an instruction-augmented multimodal sequence and feed it into $f_M$ to obtain a sequence of hidden states. Inspired by previous research~\cite{DBLP:journals/corr/abs-2308-03281, DBLP:conf/acl/WangYHYMW24, DBLP:conf/cvpr/ZhangZXLDLXZLZ25}, we use the final hidden state of the last token followed by $\ell_2$ normalization to obtain embeddings $z(x) \in \mathbb{R}^d$ in a shared joint vision-language embedding space $\mathcal{Z} \subset \mathbb{R}^d$.


\subsection{Stage 1: Pre-training the Diffusion Prior over the Joint Vision-Languange Space}
\label{sec:stage1}

The goal of the first stage is to learn a diffusion prior over fused multimodal representations in the joint vision--language space $\mathcal{Z}$. We train on a large-scale corpus of generic image-text pairs $\mathcal{D}_{\text{pair}} = \{(I_i, T_i)\}$, where each pair is mapped to a fused multimodal representation $z_0 \in \mathbb{R}^d$ by the MLLM $f_M$. Specifically, for a pair $(I, T) \in \mathcal{D}_{\text{pair}}$, we construct an instruction-augmented multimodal input $x_{\text{pair}} = (I, T)$ and obtain
\begin{equation}
    z_0 = f_M(x_{\text{pair}}),
\end{equation}
so that $z_0$ lies on the manifold of valid joint vision-language embeddings in $\mathcal{Z}$ and serves as the clean sample for diffusion.

We follow the standard DDPM formulation and apply a forward noising process to $z_0$. At timestep $n \in \{1,\dots,N\}$, a noisy fused feature is obtained in closed form as
\begin{equation}
    z_n = \sqrt{\bar{\alpha}_n}\, z_0 + \sqrt{1 - \bar{\alpha}_n}\, \epsilon,
\end{equation}
where $\epsilon \sim \mathcal{N}(0, \mathbf{I})$ and $\bar{\alpha}_n = \prod_{s=1}^{n} (1 - \beta_s)$ is defined by a fixed variance schedule $\{\beta_s\}_{s=1}^N$. The diffusion model is then trained to predict the added noise from the noisy feature $z_n$, the timestep $n$, and a textual condition derived from the paired description.

\paragraph{Pre-training Objective.}
For conditioning, we encode the text $T$ with a frozen text encoder $\mathcal{E}_{\text{text}}$ to obtain $c_T = \mathcal{E}_{\text{text}}(T)$. The denoising network $\epsilon_{\theta}$ receives $(z_n, n, c_T)$ and is optimized with an $\ell_2$ noise-prediction loss:
\begin{equation}
    \mathcal{L}_{\text{Stage1}}
    = \mathbb{E}_{(I,T) \sim \mathcal{D}_{\text{pair}},\,
                  \epsilon,\, n}
      \Bigl[
        \bigl\|
            \epsilon - \epsilon_{\theta}(z_n, n, c_T)
        \bigr\|_2^2
      \Bigr].
    \label{eq:pretrain_loss}
\end{equation}

Through this stage, $\epsilon_{\theta}$ learns a text-conditioned diffusion prior over the manifold of joint vision-language embeddings $z(x) \in \mathcal{Z}$ produced by $f_M$ from generic image-text pairs. 

\paragraph{Classifier-Free Guidance.}
\label{sec:cfg}
At sampling time, we adopt classifier-free guidance (CFG) to balance diversity and text fidelity. During pre-training, the text condition is randomly dropped with probability $p_{\mathrm{cfg}}$ to learn unconditional and conditional branches in one network. At inference, we combine the two noise predictions with guidance scale $\gamma$:
\begin{equation}
\tilde{\epsilon}_{\theta}(z_n, n, c_T)
= (1+\gamma)\,\epsilon_{\theta}(z_n, n, c_T)
 - \gamma\,\epsilon_{\theta}(z_n, n, \emptyset).
\label{eq:cfg}
\end{equation}






\subsection{Stage 2: Feature Editing over the Joint Vision-Language Space using Control-Adapter}
\label{sec:stage2}
The second stage adapts the pretrained diffusion model to composed image retrieval by editing features directly in the joint vision-language space $\mathcal{Z}$. We fine-tune on triplets $(I_r, T_{\Delta}, I_t)$, where $I_r$ is the reference image, $T_{\Delta}$ is the edit instruction, $I_t$ is the target image, and $T_t$ is the associated target description from the dataset.

For each triplet, we form two embeddings in $\mathcal{Z}$ using the MLLM $f_M$. The query-specific feature is obtained from the composed input $x_q = (I_r, T_{\Delta})$ as
\[
    z_{r,\Delta} = z(x_q) = f_M(x_q) \in \mathbb{R}^d,
\]
and the target feature is obtained from $x_t = (I_t, T_t)$ as
\[
    z_t = z(x_t) = f_M(x_t) \in \mathbb{R}^d.
\]
Together with the text condition $c_{\Delta}$, $z_{r,\Delta}$ provides query-specific conditioning and $z_t$ provides target-side supervision for diffusion-based feature editing in $\mathcal{Z}$.


\paragraph{Fine-tuning Objective.}
\label{sec:finetune_loss}

The fine-tuning objective is to train the denoiser $\epsilon_{\theta}$ to reconstruct $z_t$ from its noised version, conditioned on both the composed reference feature $z_{r,\Delta}$ and the edit text $T_{\Delta}$. Let $c_{\Delta} = \mathcal{E}_{\text{text}}(T_{\Delta})$ be the embedding of the edit text. The fine-tuning loss is formulated as:
\begin{equation}
    \mathcal{L}_{\text{Stage2}}
    = \mathbb{E}_{(I_r, T_{\Delta}, I_t),\,
                  \epsilon,\, n}
      \Bigl[
        \bigl\|
            \epsilon - \epsilon_{\theta}\bigl(z_n, n, z_{r,\Delta}, c_{\Delta}\bigr)
        \bigr\|_2^2
      \Bigr],
    \label{eq:finetune_loss}
\end{equation}


\paragraph{Control-Adapter for query-specific conditioning}
Let $\epsilon_{\theta}$ denote the denoising network operating in $\mathcal{Z}$, whose encoder is composed of $L$ blocks $\{F_l(\cdot;\Theta_l)\}_{l=1}^{L}$. To enable precise conditioning on the query-specific feature while preserving the original generative prior, we introduce a Control-Adapter branch that augments each block with a trainable copy and zero-initialized linear layers. For each DiT encoder block $F(\cdot;\Theta)$, we instantiate a control block $F_c(\cdot;\Theta_c)$ initialized from $\Theta$, together with two types of zero linear layers $Z_1(\cdot;\Theta_{z1})$ and $Z_2(\cdot;\Theta_{z2})$.

Denote by $\mathbf{h}^{enc}_{l-1}$ the input to the $l$-th DiT encoder block and by $\mathbf{h}^{enc}_{c,l-1}$ the input to the $l$-th control block. For the output of the Control-Adapter $\mathbf{y}_c = \{\mathbf{y}_{c,1}, \mathbf{y}_{c,2}, ..., \mathbf{y}_{c,l}, ..., \mathbf{y}_{c,L}\}$, where $\mathbf{y}_{c,l}$ is the $l$-th block output of Control Adapter, which becomes the input for the $(L-l+1)$-th decoder block of DiT:
\begin{equation}
    \mathbf{y}_{c,l} = F(\mathbf{h}^{enc}_{l-1}; \Theta_l) + Z_2\bigl(F_c(\mathbf{h}^{enc}_{c,l-1};\Theta_{c,l});\Theta_{z2,l}\bigr)
\end{equation}
where $\mathbf{h}^{enc}_{c,l-1} = \mathbf{h}^{enc}_{l-1} + Z_1(z_{r,\Delta};\Theta_{z1})$ for $l = 1$, and $\mathbf{h}^{enc}_{c,l-1} = \mathbf{y}_{c,l-1}$ for $l > 1$. Both $Z_1$ and $Z_2$ are initialized with zero weights and biases. At the beginning of training, their outputs are exactly zero, so that $\mathbf{y}_{c,l} = F(\mathbf{h}^{enc}_{l-1}; \Theta_l)$ and the Control-Adapter does not perturb the original denoising behavior. As training proceeds, the control branch gradually learns how to modulate intermediate features using $z_{r,\Delta}$, while the DiT parameters remain frozen.

For the decoder, we inject the control signal by residual addition. Let $\tilde{\mathbf{h}}^{dec}_l$ denote the output of the $l$-th decoder block. We feed $\tilde{\mathbf{h}}^{dec}_l + \delta \mathbf{y}_{c,L-l}$ into the ($l+1$)-th decoder block, while the parameters of DiT encoder and decoder remain frozen. In this way, the pretrained model provides a strong generative prior, and the Control-Adapter learns a lightweight, query-conditioned residual that steers the feature editing process in $\mathcal{Z}$ without causing catastrophic forgetting.

\subsection{Inference and Retrieval}
\label{sec:inference}

\paragraph{Query and Condition Generation.}
\label{sec:inference_query}
At inference time, given a composed query $(I_r, T_{\Delta})$, we first compute its fused representation and the corresponding textual condition in the joint vision-language space $\mathcal{Z}$. The query-specific embedding is obtained from the multimodal input
\[
    x_q = (I_r, T_{\Delta}), \quad
    z_{r,\Delta} = z(x_q) = f_M(x_q) \in \mathbb{R}^d.
\]

To provide richer textual guidance than the raw edit instruction, we further use the same MLLM $f_M$ to infer a detailed target-oriented description from the query. Specifically, we construct an instruction-augmented input
\[
    x_{\text{cond}} = (\texttt{[INST]}, I_r, T_{\Delta}),
\]
where \texttt{[INST]} denotes a textual instruction for describing the desired target, and obtain $\tilde{T} = f_M(x_{\text{cond}}).$
The inferred description is then encoded by the frozen text encoder to produce $c_{\tilde{T}} = \mathcal{E}_{\text{text}}(\tilde{T}).$

\paragraph{Target Feature Sampling.}
\label{sec:inference_sampling}
We then run the learned reverse diffusion process in $\mathcal{Z}$ to generate a single target feature hypothesis. Starting from an initial noise vector $\hat{z}_N \sim \mathcal{N}(0, \mathbf{I})$,
we iteratively denoise for timesteps $n = N,\dots,1$ using the denoiser $\epsilon_{\theta}$:
\begin{equation}
    \hat{z}_{n-1}
    = \frac{1}{\sqrt{\alpha_n}}
      \left(
        \hat{z}_n
        - \frac{1 - \alpha_n}{\sqrt{1 - \bar{\alpha}_n}}\,
          \epsilon_{\theta}\bigl(\hat{z}_n, n, z_{r,\Delta}, c_{\tilde{T}}\bigr)
      \right)
      + \sigma_n \mathbf{w}_n,
    \label{eq:sampling}
\end{equation}
where $\mathbf{w}_n \sim \mathcal{N}(0,\mathbf{I})$ and $\sigma_n$ is the noise standard deviation at timestep $n$. After $N$ steps, we obtain the final target feature hypothesis $\hat{z}_t = \hat{z}_0 \in \mathcal{Z}.$

\paragraph{Retrieval and Ranking.}
\label{sec:inference_ranking}
For retrieval, each gallery image $I_g$ in the database $\mathcal{D}_{\text{gallery}}$ is embedded into the same joint space using the MLLM $f_M$:
\[
    z_g = f_M(I_g) \in \mathbb{R}^d.
\]
The relevance score $s(I_g)$ for $I_g$ is computed as the cosine similarity between $\hat{z}_t$ and $z_g$:
\begin{equation}
    s(I_g)
    = \frac{\hat{z}_t^{\top} z_g}
           {\|\hat{z}_t\|_2\, \|z_g\|_2}.
    \label{eq:retrieval_score}
\end{equation}
Gallery images are ranked according to $s(I_g)$, and the top-ranked ones are returned as retrieval results. Since both the generated target feature $\hat{z}_t$ and the gallery embeddings $z_g$ reside in the shared joint vision--language space $\mathcal{Z}$, retrieval is conducted entirely in a unified representation space while leveraging rich target-oriented textual guidance at inference time.


\section{Experiments}
\label{sec:experiments}
In this section We conduct extensive experiments to systematically evaluate our proposed Fusion-Diff framework and validate its effectiveness in addressing the core challenges of Zero-Shot Composed Image Retrieval (ZS-CIR). 

\subsection{Experimental Settings}
\paragraph{Datasets and Baselines.}
We evaluate Fusion-Diff on three standard ZS-CIR benchmarks and an additional real-world dataset introduced in this work, comparing with several state-of-the-art methods. CIRR~\cite{DBLP:conf/iccv/0002OTG21} is built from real-life images which contains 21,552 images organized into groups of semantically related scenes and triplets (reference, textual modification, target). FashionIQ~\cite{DBLP:conf/cvpr/WuGGARGF21} is a fashion retrieval benchmark containing 77,684 images of clothing items from three categories (Dress, Shirt, Toptee) and 30,134 triplets. CIRCO~\cite{DBLP:conf/iccv/BaldratiA0B23} is a compositional retrieval benchmark on MS-COCO, featuring natural-language instructions about fine-grained attribute, layout, and content changes, with 1,020 test queries, 4,624 annotated target images, and a fixed COCO image gallery. We compare Fusion-Diff against representative state-of-the-art ZS-CIR methods, including Pic2Word~\cite{DBLP:conf/cvpr/SaitoSZLLSP23a}, SEARLE~\cite{DBLP:conf/iccv/BaldratiA0B23} and LinCIR~\cite{DBLP:conf/cvpr/GuCKKY24} under CLIP backbone ViT-G. We further include training-free LLM/MLLM reasoning approaches (CIReVL~\cite{DBLP:conf/iclr/KarthikRMA24}, LDRE~\cite{DBLP:conf/sigir/YangXQDX24}, OSrCIR~\cite{DBLP:conf/cvpr/TangZQYG0LR0025}) and generative baselines (CIG~\cite{DBLP:conf/cvpr/WangABL25}, IP-CIR~\cite{DBLP:conf/cvpr/LiMY25}) 

\paragraph{Evaluation Metrics.}
For CIRR, we follow the original benchmark protocol and report $\mathrm{Recall}@k$ as the primary metric, where $k \in \{1, 5, 10, 50\}$. We further evaluate performance in the subset setting using $\mathrm{RecallSubset}@k$ with $k \in \{1, 2, 3\}$. On FashionIQ, we similarly adopt $\mathrm{Recall}@k$ for evaluation, with $k \in \{10, 50\}$. For CIRCO, since each query is associated with multiple target images, we use mean Average Precision at rank $k$ ($\mathrm{mAP}@k$), where $k \in \{5, 10, 25, 50\}$ denotes the size of the top-ranked retrieval list.

\paragraph{Implementation Details.}
We implement our framework in PyTorch, training on 8 NVIDIA A100 GPUs. We adopt a T5 encoder for text processing and Qwen2.5VL-72B-Instruct~\cite{DBLP:journals/corr/abs-2502-13923} as the MLLM $f_M$. Further details on the model architecture and hyperparameters are available in the Appendix.

\begin{table*}[t]
\centering
\setlength{\tabcolsep}{4.2pt}
\renewcommand{\arraystretch}{1.15}
\begin{threeparttable}
\caption{\textbf{Comparison on CIRCO and CIRR Test Data.}
Bold and \underline{underline} denote the best and second-best, respectively.}
\label{tab:circo-cirr}
\small
\begin{tabular}{l|l|cccc d ccccccc}
\toprule
\multicolumn{2}{c|}{\textbf{CIRCO + CIRR $\rightarrow$}} &
\multicolumn{4}{c|}{\textbf{CIRCO}} &
\multicolumn{7}{c}{\textbf{CIRR}} \\
\cmidrule(lr){3-6}\cmidrule(lr){7-13}
\multicolumn{2}{c|}{\multirow{2}{*}{\textbf{Backbone\;\;|\;\;Method}}} &
\multicolumn{4}{c|}{\textbf{mAP@k}} &
\multicolumn{4}{c|}{\textbf{Recall@k}} &
\multicolumn{3}{c}{\textbf{Recall$_{\text{subset}}$@k}} \\
\multicolumn{2}{c|}{} & $k{=}5$ & $k{=}10$ & $k{=}25$ & $k{=}50$ &
$k{=}1$ & $k{=}5$ & $k{=}10$ & $k{=}50$ & $k{=}1$ & $k{=}2$ & $k{=}3$ \\
\midrule
\multirow{3}{*}{ViT-B/32}
  & Image-only             & 1.34 & 1.60 & 2.12 & 2.41 & 6.89 & 22.99 & 33.68 & 59.23 & 21.04 & 41.04 & 60.31   \\
  & Text-only               & 2.56 & 2.67 & 2.98 & 3.18 & 21.81 & 45.22 & 57.42 & 81.01 & 62.24 & 81.13 & 90.70 \\
  & Image + Text               & 2.65 & 3.25 & 4.14 & 4.54 & 11.71 & 35.06 & 48.94 & 77.49 & 32.77 & 56.89 & 74.96 \\

\midrule
\multirow{2}{*}{ViT-L/14}
  & Pic2Word             &  8.72 &  9.51 & 10.64 & 11.29 & 23.90 & 51.70 & 65.30 & 87.80 & 53.76 & 74.46 & 87.07   \\
  & SEARLE-XL               & 11.68 & 12.73 & 14.33 & 15.12 & 24.24 & 52.48 & 66.29 & 88.84 & 53.76 & 75.01 & 88.19 \\
  
\midrule
\multirow{8}{*}{ViT-G/14}
  & LinCIR               & 19.71 & 21.01 & 23.13 & 24.18 & 35.25 & 64.72 & 76.05 & - & 63.35 & 82.22 & 91.98 \\
  & LinCIR + CIG-XL        & 20.64 & 21.90 & 24.04 & 25.20 & 34.43 & 64.51 & 76.12 & 93.54 & 62.24 & 81.35 & 91.28 \\
  & LinCIR + IP-CIR   & 25.70 & 26.64 & 29.09 & 30.13 & 35.37 & 64.70 & 76.15 & 93.71 & 62.58 & 81.74 & 91.35 \\
  & CompoDiff   & 15.33 & 17.71 & 19.45 & 21.01 & 26.71 & 55.14 & 74.52 & 92.01 & 64.54 & 82.39 & 91.81 \\
  & CIReVL            & 26.77 & 27.59 & 29.96 & 31.03 & 34.65 & 64.29 & 75.06 & 91.66 & 67.95 & 84.87 & 93.21 \\
  & LDRE               & 31.12 & 32.24 & 34.95 & 36.03 & 36.15 & 66.39 & 77.25 & 93.95 & 68.82 & 85.66 & 93.76 \\
  & LDRE + IP-CIR   & \underline{32.75} & \underline{34.26} & \underline{36.86} & \underline{38.03} & \underline{39.25} & \underline{70.07} & \underline{80.00} & \underline{94.89} & \best{69.95} & \underline{86.87} & \underline{94.22} \\
  & OSrCIR               & 30.47 & 31.14 & 35.03 & 36.59 & 37.26 & 67.25 & 77.33 & - & 69.22 & 85.28 & 93.55 \\
  \midrule
\multirow{1}{*}{}
  & \textbf{Fusion-Diff}      & \best{35.26} & \best{36.58} & \best{39.17} & \best{40.37} &
                           \best{40.00} & \best{70.27} & \best{81.28} & \best{95.33} & \underline{69.83} & \best{86.90} & \best{94.24} \\
\bottomrule
\end{tabular}

\end{threeparttable}
\end{table*}

\begin{table*}[t]
\centering
\setlength{\tabcolsep}{5pt}
\renewcommand{\arraystretch}{1.15}
\begin{threeparttable}
\caption{\textbf{Comparison on FashionIQ Validation Data.}
Bold and \underline{underline} denote the best and second-best, respectively.}
\label{tab:fashioniq}
\small
\begin{tabular}{l|l|cc|cc|cc|cc}
\toprule
\multicolumn{2}{c|}{\textbf{Fashion-IQ $\rightarrow$}} &
\multicolumn{2}{c|}{\textbf{Shirt}} &
\multicolumn{2}{c|}{\textbf{Dress}} &
\multicolumn{2}{c|}{\textbf{Toptee}} &
\multicolumn{2}{c}{\textbf{Average}} \\
\cmidrule(lr){3-4}\cmidrule(lr){5-6}\cmidrule(lr){7-8}\cmidrule(lr){9-10}
\multicolumn{2}{c|}{\textbf{Backbone\;\;|\;\;Method}} &
\textbf{R@10} & \textbf{R@50} &
\textbf{R@10} & \textbf{R@50} &
\textbf{R@10} & \textbf{R@50} &
\textbf{R@10} & \textbf{R@50} \\
\midrule
\multirow{3}{*}{ViT-B/32}
 & Image-only          & 6.92 & 14.23 & 4.46 & 12.19 & 6.32 & 13.77 & 5.90 & 13.37 \\
 & Text-only          & 19.87 & 34.99 & 15.42 & 35.05 & 20.81 & 40.49 & 18.70 & 36.84 \\
 & Image + Text   & 13.44 & 26.25 & 13.83 & 30.88 & 17.08 & 31.67 & 14.78 & 29.60 \\
\midrule
\multirow{2}{*}{ViT-L/14}
 & Pic2Word        & 26.20 & 43.60 & 20.00 & 40.20 & 27.90 & 47.40 & 24.70 & 43.70 \\
 & SEARLE          & 26.89 & 45.58 & 20.48 & 43.13 & 29.32 & 49.97 & 25.56 & 46.23 \\
\midrule
\multirow{7}{*}{ViT-G/14}
& LinCIR          & 46.76 & 65.11 & 38.08 & 60.88 & 50.48 & 71.09 & 45.11 & 65.69 \\
& LinCIR + CIG-XL          & 47.35 & \underline{66.68} & \underline{39.71} & 60.93 & \underline{50.69} & \underline{71.39} & \underline{45.92} & \underline{66.34} \\
& LinCIR + IP-CIR          & \underline{48.04} & \underline{66.68} & 39.02 & \underline{61.03} & 50.18 & 71.14 & 45.74 & 66.28 \\
& CompoDiff       & 41.31 & 55.17 & 37.78 & 49.10 & 44.26 & 56.41 & 39.02 & 51.71 \\
& CIReVL      & 34.01 & 51.92 & 27.56 & 50.04 & 36.29 & 56.63 & 32.62 & 52.86 \\
& LDRE   & 35.94 & 58.58 & 26.11 & 51.12 & 35.42 & 56.67 & 32.49 & 55.46 \\
& OSrCIR & 38.65 & 54.71 & 33.02 & 54.78 & 41.04 & 61.83 & 37.57 & 57.11 \\
\midrule
\multirow{1}{*}{}
  & \textbf{Fusion-Diff}      & \best{48.23} & \best{68.45} & \best{41.15} & \best{63.76} &
                           \best{51.30} & \best{71.55} & \best{46.89} & \best{67.92} \\

\bottomrule
\end{tabular}

\end{threeparttable}
\end{table*}

\subsection{Quantitative Results on ZS-CIR Benchmarks}
\label{sec:quant}
\paragraph{CIRCO.}
We present the experimental results on the CIRCO test dataset in the left half of Table~\ref{tab:circo-cirr}. Across all four cutoffs, Fusion-Diff attains the best $\mathrm{mAP}@k$ (e.g., $40.37$ at $k{=}50$), improving over the strongest prior baseline (\textit{LDRE+IP-CIR}) by +2.34 absolute at $\mathrm{mAP}@50$ (+6.1\%), +2.31 at $\mathrm{mAP}@25$, +2.32 at $\mathrm{mAP}@10$, and +2.51 at $\mathrm{mAP}@5$. These gains hold against both training-free reasoning (CIReVL/LDRE/OSrCIR) and generative pipelines (LinCIR{+}CIG/IP-CIR, CompoDiff), indicating that editing fused features via diffusion is more robust than caption-only reasoning or pixel-space pseudo-targets on the compositional COCO benchmark.

\paragraph{CIRR.}
We present the experimental results on the CIRR test dataset in the right half of Table~\ref{tab:circo-cirr}. On global retrieval, Fusion-Diff establishes new state of the art at all $k$ (e.g., $\mathrm{R}@1{=}40.00$, $\mathrm{R}@10{=}81.28$), surpassing LDRE+IP-CIR by +0.75 at $\mathrm{R}@1$, +1.28 at $\mathrm{R}@10$, and +0.44 at $\mathrm{R}@50$. On group-based evaluation, our method is \textit{best} on $\mathrm{R}_{\text{subset}}@2/@3$ and \textit{second} on $\mathrm{R}_{\text{subset}}@1$ (69.83 vs.\ 69.95 for LDRE+IP-CIR), narrowing the gap to the strongest one-stage CoT pipeline (OSrCIR) while outperforming two-stage reasoning (CIReVL). The pattern supports our design choice of operating directly in the joint vision–language space rather than relying solely on language-side edits.

\paragraph{FashionIQ.}
In Table~\ref{tab:fashioniq}, Fusion-Diff achieves the best results on all three categories and both metrics. For the averaged score, we obtain $\mathrm{R}@10{=}46.89$ and $\mathrm{R}@50{=}67.92$, improving over the strongest prior systems (LinCIR variants with/without generative auxiliaries) by +0.97 and +1.58 absolute, respectively (Table~\ref{tab:fashioniq}). Per-category gains are consistent—e.g., on \textit{Dress} our method reaches 41.15/63.76 at R@10/R@50 (vs.\ 39.71/61.03 for LinCIR{+}CIG/IP-CIR), and on \textit{Shirt} we set new highs at 48.23/68.45. These results indicate that our diffusion-based feature editing preserves fine-grained garment identity while applying attribute changes more faithfully than caption-only reasoning or textual-inversion baselines.

\begin{figure}[htbp]
\centering
\vspace{-1ex}
\includegraphics[width=1.0\linewidth]{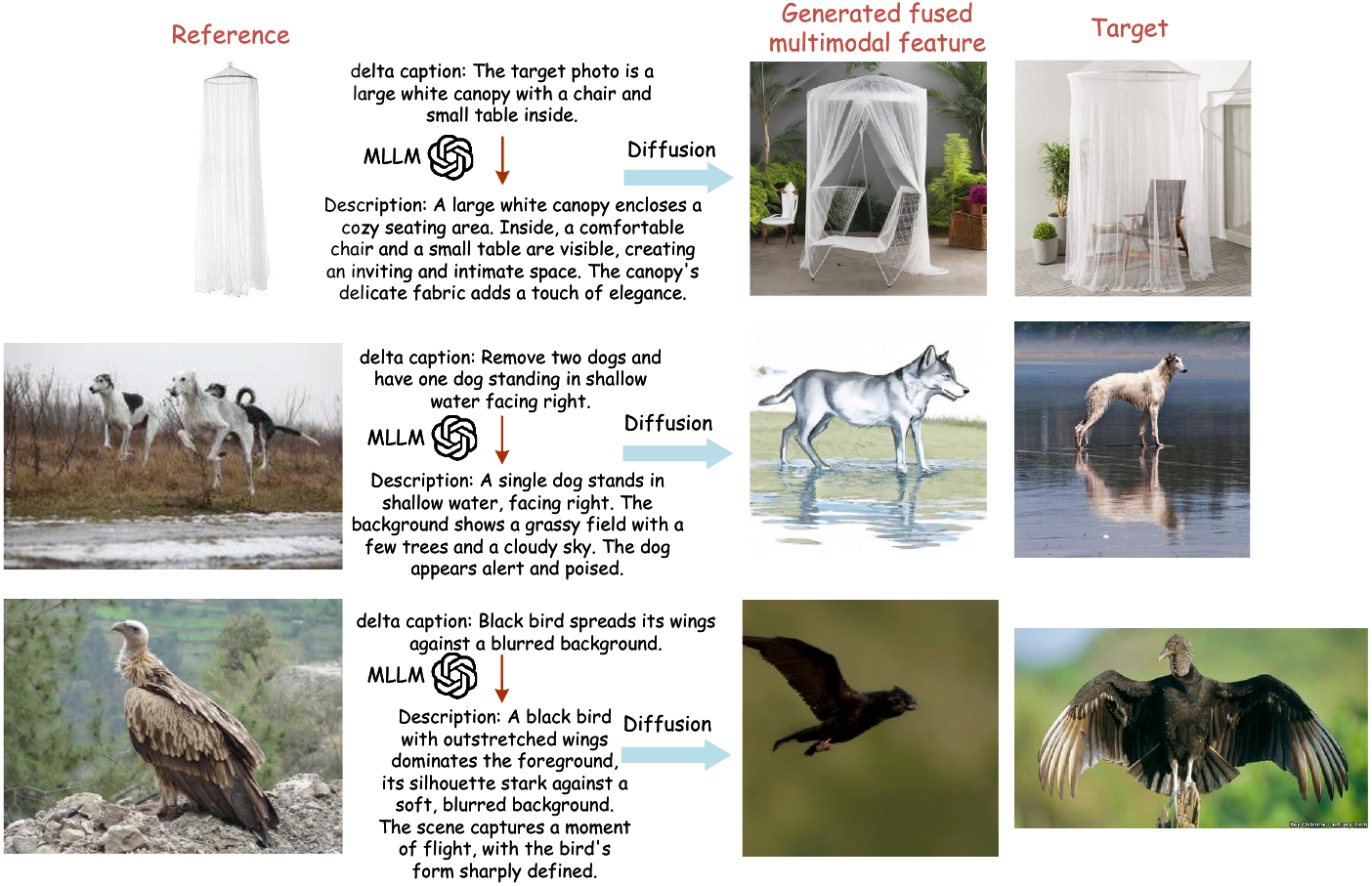}
\caption{Visualization results on CIRR test set.}
\centering
 \label{CIRRstudy}
 \vspace{-3ex}
\end{figure}

\vspace{7em}

\begin{figure}[htbp]
\centering
\vspace{-1ex}
\includegraphics[width=1.0\linewidth]{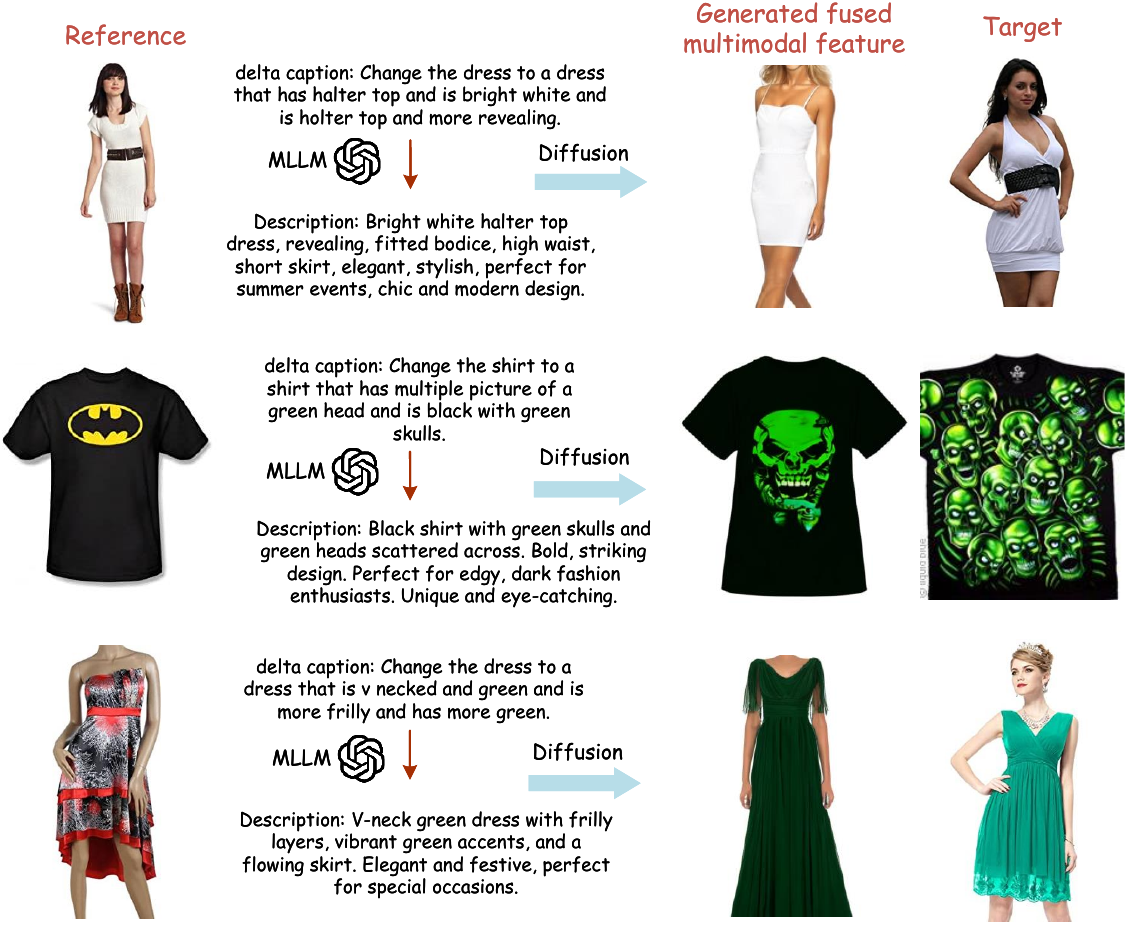}
\caption{Visualization results on FashionIQ validation set.}
\centering
 \label{FashionIQstudy}
 \vspace{-3ex}
\end{figure}

\subsection{Visualization of Joint Multimodal Features}

To validate the effectiveness and interpretability of our method, we fine-tuned a decoder based on SANA 1.5~\cite{DBLP:journals/corr/abs-2501-18427} to reconstruct images from the generated multimodal features for retrieval.
Figure~\ref{CIRRstudy}~ shows the visualization results on CIRR. We observe that our method successfully interprets challenging compositional modifications, retrieving targets that require both preserving the reference image's core identity and accurately applying fine-grained semantic edits. Similarly, Figure~\ref{FashionIQstudy} demonstrates our model's precision on FashionIQ, adeptly handling attribute-specific edits (e.g., color, texture, or sleeve length). These results validate our framework's ability to perform effective retrieval in the joint multimodal space, overcoming the modality gap seen in prior cross-modal methods.

\begin{table*}[t]
\centering
\setlength{\tabcolsep}{6pt}
\renewcommand{\arraystretch}{1.15}
\begin{threeparttable}
\caption{Ablations on FashionIQ and CIRCO.
(A) $z_{r,\Delta}$-only: Uses the fused feature with no diffusion.
(B) Stage-I-only: Uses text-conditioned diffusion prior \emph{without} image-aware control.
(C) Full Model (Fusion-Diff): Uses the full pipeline with \emph{both} the Control-Adapter and text-side conditioning.
}
\label{tab:ablation-fiq-circo}
\small
\begin{tabular}{l|cc|cccc}
\toprule
\multirow{2}{*}{\textbf{Variant}} 
& \multicolumn{2}{c|}{\textbf{FashionIQ (Avg.)} $\uparrow$} 
& \multicolumn{4}{c}{\textbf{CIRCO} $\uparrow$} \\
& \textbf{R@10} & \textbf{R@50} 
& \textbf{mAP@5} & \textbf{mAP@10} & \textbf{mAP@25} & \textbf{mAP@50} \\
\midrule
(A) $z_{r,\Delta}$-only (no diffusion) 
& \underline{40.57} & \underline{61.22} 
& \underline{14.91} & \underline{16.44} & \underline{18.57} & \underline{19.50} \\

(B) Stage-I-only (text-conditioned prior) 
& 33.12 & 54.08 
& 13.02 &  14.16 & 16.62 & 17.87    \\

(C) \textbf{Full Model (Fusion-Diff)} 
& \textbf{46.89} & \textbf{67.92} 
& \textbf{35.26} & \textbf{36.58} & \textbf{39.17} & \textbf{40.37} \\
\bottomrule
\end{tabular}
\begin{tablenotes}[flushleft]\footnotesize
\item \textbf{Avg.} denotes the arithmetic mean over the three FashionIQ categories (Dress, Shirt, Toptee).
\end{tablenotes}
\end{threeparttable}
\end{table*}

\subsection{Ablation Studies}
\label{sec:ablation}
We conduct controlled ablations on FashionIQ and CIRCO to disentangle and validate the contributions of our core components, as shown in Table~\ref{tab:ablation-fiq-circo}. We analyze three main variants: (A) a baseline using only the fused feature without diffusion, (B) a diffusion model conditioned \emph{only} on text, and (C) our complete Fusion-Diff framework.

\paragraph{Settings.}
\textbf{(A) $z_{r,\Delta}$-only retrieval (no diffusion).}
Given a composed query $(I_r, T_{\Delta})$, we embed it once to obtain $z_{r,\Delta}$ and rank gallery items by cosine similarity $s(I_g)=\langle z_{r,\Delta}, z_g\rangle$, with no diffusion sampling or editing. This baseline quantifies how far a strong fused representation can go without any generative refinement.

\textbf{(B) Stage-I-only (text-conditioned prior).}
We use only the pretrained text-conditioned diffusion prior from Stage~I to generate a single target feature hypothesis $\hat z_t$ \emph{without} Stage~II editing. Concretely, we drop $z_{r,\Delta}$ and condition the reverse process with the text side only (using the MLLM-inferred $\tilde{T}$) to obtain $\hat z_t$, which is then used for retrieval. This isolates the effect of a generic text-to-feature generative prior.

\textbf{(C) Full Model (Fusion-Diff).}
This is our full proposed framework, which executes Stage~II feature editing using both the text-side conditioning (driven by $\tilde{T}$) and the image-aware Control-Adapter (driven by $z_{r,\Delta}$). This variant matches the main results reported in Tables~\ref{tab:circo-cirr} and \ref{tab:fashioniq}.

\paragraph{Discussion.}
The results in Table~\ref{tab:ablation-fiq-circo} validate our design:

\begin{itemize}
    \item Necessity of Diffusion Refinement.
    Variant (A) ($z_{r,\Delta}$-only) establishes a strong baseline (e.g., 40.57 R@10 on FashionIQ), indicating a high-quality initial query embedding. However, our Full Model (C) dramatically improves upon this (46.89 R@10), proving that the generative refinement provided by our diffusion process is crucial for achieving top-tier performance.

    \item Necessity of Control-Adapter. 
    Variant (B) (text-only diffusion) performs very poorly (33.12 R@10 on FashionIQ, 13.02 mAP@5 on CIRCO), scoring significantly \emph{worse} than the non-diffusion baseline (A). This highlights a critical failure mode: text-conditioning alone is insufficient and fails to preserve the reference image's core visual identity.

    \item Synergy of Dual-Conditioning.
    Our Full Model (C) succeeds by combining both pathways. By comparing (C) against (A) and (B), we conclude that \emph{both} components are essential. The Control-Adapter (missing in B) is vital for preserving visual identity, while the text-conditioning (analyzed in Sec.~\ref{sec:param_sensitivity}) is vital for guiding the feature towards the correct semantic edit. This dual-conditioning synergy is the key to our state-of-the-art results.
\end{itemize}

\begin{figure}[htbp]
  \centering
  \vspace{-1ex}
  \begin{subfigure}[b]{0.48\linewidth}
    \centering
    \includegraphics[width=\linewidth]{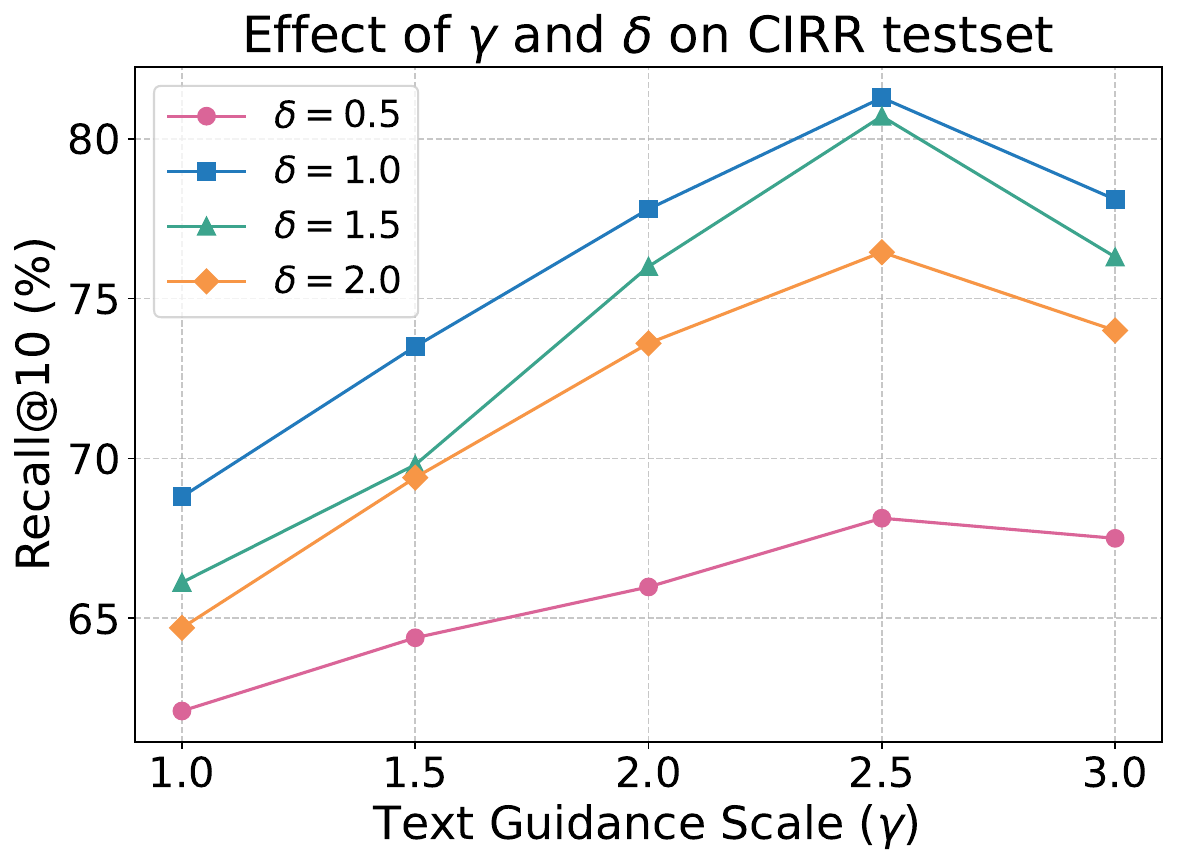}
  \end{subfigure}
  \hfill
  \begin{subfigure}[b]{0.51\linewidth}
    \centering
    \includegraphics[width=\linewidth]{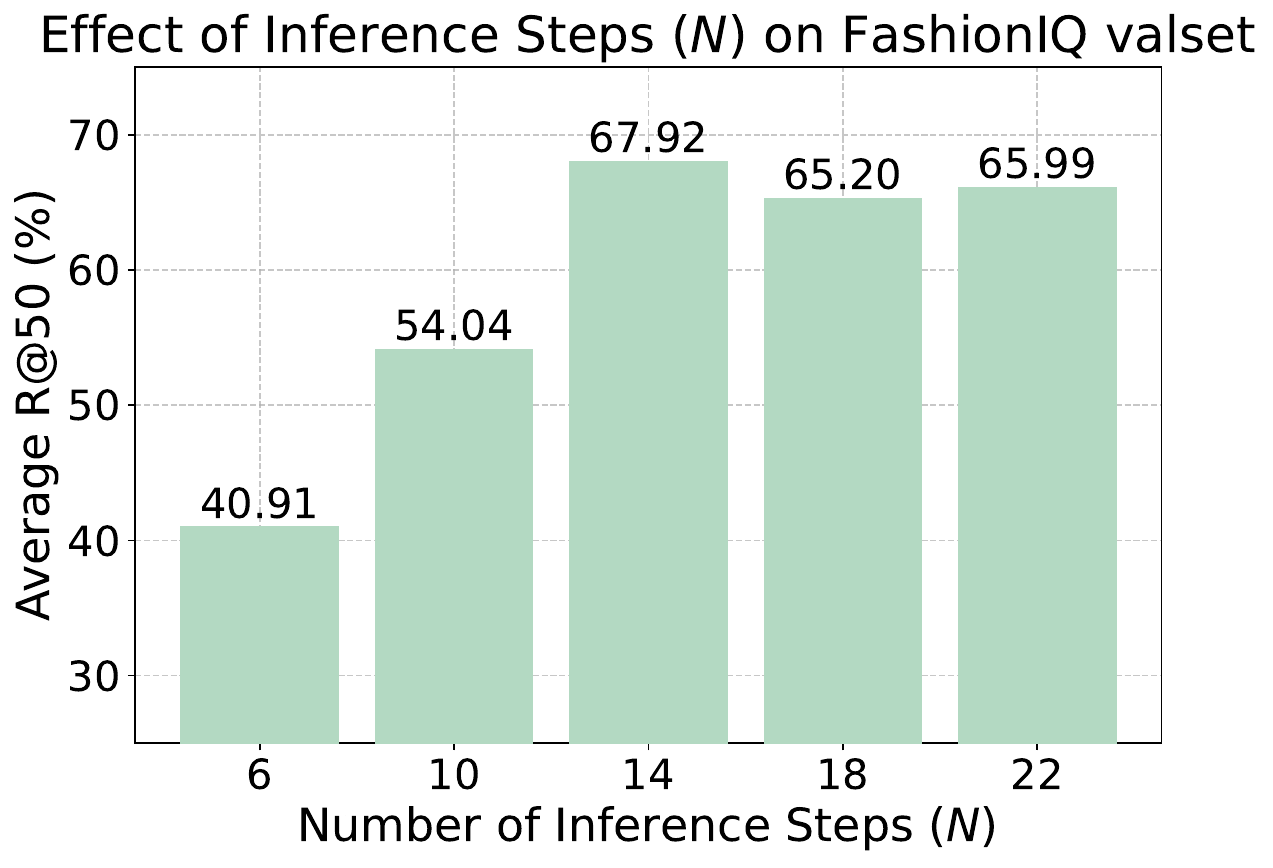}
  \end{subfigure}  
  \caption{Parameter sensitivity analysis on CIRR test set and FashionIQ validation set.}
  \label{fig:sensitivity1}
  \vspace{-1ex}
\end{figure}

\subsection{Parameter Sensitivity Analysis}
\label{sec:param_sensitivity}
We study the sensitivity of three key hyperparameters, with results shown in Figure~\ref{fig:sensitivity1}: the text-guidance scale $\gamma$ and Control-Adapter strength $\delta$ (analyzed on CIRR R@10), and the number of denoising steps $N$ (analyzed on FashionIQ Avg. R@50).

\paragraph{Effect of $\gamma$ and $\delta$.}
As shown in Figure~\ref{fig:sensitivity1} (left), performance is sensitive to the balance between text guidance ($\gamma$) and visual control ($\delta$). We observe a clear peak at $(\gamma=2.5, \delta=1.0)$, achieving 81.28\% R@10 on CIRR. Performance degrades if either guidance is too weak (e.g., $\delta=0.5$) or too strong (e.g., $\gamma > 2.5$), confirming the need for a balanced interplay to apply the edit without overriding the reference image.

\paragraph{Effect of $N$.}
Figure~\ref{fig:sensitivity1} (right) shows the impact of denoising steps $N$ on FashionIQ Avg. R@50. Performance is poor with too few steps (40.91\% at $N=6$) but rises sharply to a peak of 67.92\% at $N=14$. Beyond this, returns diminish and performance slightly degrades (e.g., at $N=18, 22$) while latency increases. We thus select $N=14$ as the optimal accuracy–efficiency trade-off.

\section{Conclusion and Limitations}
\label{sec:conclusion}
We propose Fusion-Diff, a novel generative framework that reframes zero-shot composed image retrieval (ZS-CIR) as multimodal-to-multimodal matching. Our Control-Adapter architecture mitigates data inefficiency through lightweight, parameter-efficient fine-tuning, while the joint vision-language space representation bridges the modality gap. Experiments on CIRR, FashionIQ, and CIRCO demonstrate superior performance over existing methods. Nevertheless, standard diffusion sampling introduces additional computational overhead; future work will explore acceleration techniques.
\maketitlesupplementary

\section{Zero-Shot Protocol and Data Settings}

In all experiments, we strictly adhere to the zero-shot composed image retrieval (ZS-CIR) protocol. Our model is never trained on any triplets, images, or annotations from the evaluation benchmarks (CIRR~\cite{DBLP:conf/iccv/0002OTG21}, FashionIQ~\cite{DBLP:conf/cvpr/WuGGARGF21}, and CIRCO~\cite{DBLP:conf/iccv/BaldratiA0B23}). All training data are derived from public datasets with domains disjoint from the test sets, ensuring that evaluation images and captions remain completely unseen during the training phase.

\subsection{VL Diffusion Prior Pre-Training}
To learn a robust generative prior, we construct a large-scale corpus, $\mathcal{D}_{\text{pair}}$, comprising 33 million image-text pairs collected from public datasets distinct from the downstream benchmarks. Following the data construction recipe of BLIP3-o~\cite{DBLP:journals/corr/abs-2505-09568}, we aggregate images from open-source datasets including CC12M~\cite{DBLP:conf/cvpr/ChangpinyoSDS21}, SA-1B~\cite{DBLP:conf/iccv/KirillovMRMRGXW23}, and JourneyDB~\cite{DBLP:conf/nips/SunPGLDWZZQWDQW23}. To ensure the model generalizes across varying text granularities, we implement a mixed-length captioning strategy:

\begin{itemize}
    \item \textbf{Long-Text Subset (28M):} To ensure semantically rich supervision, this subset comprises detailed synthetic captions generated by Qwen2.5-VL-7B-Instruct~\cite{DBLP:journals/corr/abs-2502-13923}, with an average length of approximately 120 tokens.
    \item \textbf{Short-Text Subset (5M):} To mitigate overfitting to verbose descriptions and maintain alignment with concise queries, this subset consists of original short captions (avg. 20 tokens) sourced from the CC12M dataset.
\end{itemize}

For every pair $(I, T) \in \mathcal{D}_{\text{pair}}$, we feed the image and text into the frozen MLLM $f_M$ to obtain a fused vision--language (VL) representation $z_0 = f_M(I, T)$ within the joint embedding space. Stage 1 trains the text-conditioned diffusion model to estimate the distribution of these fused features. This pre-training strategy offers three key advantages:
\begin{enumerate}
    \item \textbf{Geometric Alignment:} It aligns the diffusion prior with the intrinsic geometry of the joint VL space used for retrieval.
    \item \textbf{Robustness:} It enhances robustness to varying text lengths and styles, as the model is exposed to a diverse range of caption distributions (from the 28M detailed descriptions to the 5M concise captions).
    \item \textbf{Generalization:} It establishes a generic, dataset-agnostic prior that facilitates effective transfer to the composed image retrieval task using limited synthetic data.
\end{enumerate}

\subsection{Synthetic Editing Alignment}
In the second stage, we leverage the HQEdit dataset~\cite{DBLP:conf/iclr/HuiYZSWWXZ25} to instill query-dependent editing capabilities into the joint VL space via the Control-Adapter. HQEdit contains approximately $200\mathrm{K}$ synthetic triplets $(I_r, T_{\Delta}, I_t)$, generated to provide high-quality instruction-following examples while remaining disjoint from real-world CIR benchmarks. This stage fine-tunes the Control-Adapter to modulate the pre-trained prior based on the reference image $I_r$ and edit instruction $T_{\Delta}$.

\subsection{Inference-Time Target Description Generation}
During inference, our framework requires a detailed textual description of the target image to guide the diffusion process in the VL space. Since ground-truth target captions are unavailable in the ZS-CIR setting, we employ the MLLM $f_M$ to hallucinate a pseudo-target description $\tilde{T}$ based on the reference image $I_r$ and the edit instruction $T_{\Delta}$.

We construct specific instruction prompts (denoted as \texttt{INST}) for each benchmark to optimize the MLLM's generation quality. The prompts are formulated as follows:

To optimize the MLLM's generation quality and ensure semantic consistency across all benchmarks (CIRCO, CIRR, and FashionIQ), we adopt a unified instruction template. This template explicitly enforces a constraint to preserve the core visual identity while applying the modification. The prompt is formulated as follows:

\begin{quote}
    \textit{``$<$image$>$ Change the image into a new one according to the requirement of the editing prompt `$\{T_{\Delta}\}$' with a shared concept `$\{\text{shared\_concept}\}$' remaining unchanged. Objectively describe the new image after editing.''}
\end{quote}

In this template, the shared concept is instantiated with the coarse category of the reference image (e.g., ``dog'' for CIRR or ``dress'' for FashionIQ) to anchor the MLLM's reasoning. The generated description $\tilde{T}$ is then embedded and used alongside the reference image $I_r$ to sample the final target feature $z_t$ via the diffusion prior.

\section{Implementation Details of Fusion-Diff}

\subsection*{Network Architecture of the Diffusion Prior}

\begin{figure}[t]
    \centering
    \includegraphics[width=\linewidth]{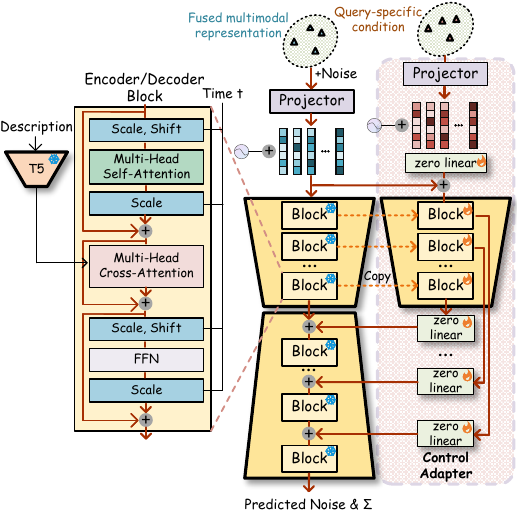}
    \caption{Architecture of the Fusion-Diff Diffusion Prior. The model processes fused VL embeddings as pseudo-spatial feature maps, modulated by timestep and text conditions via a transformer backbone.}
    \label{fig:appendix_arch_final}
\end{figure}

We formulate the diffusion prior as a conditional Diffusion Transformer (DiT)~\cite{DBLP:conf/iccv/PeeblesX23} operating directly within the joint vision--language (VL) manifold. Let $f_M$ denote the MLLM. For an input image--text pair, $f_M$ yields a fused representation $z_0 \in \mathbb{R}^{d_{\mathrm{VL}}}$, where $d_{\mathrm{VL}} = 1536$. We model $z_0 \sim p_{\mathrm{data}}(z_0)$ as the data variable, eschewing pixel-space decoding or VAE latent compression. The overall architecture is illustrated in Figure~\ref{fig:appendix_arch_final}.

\paragraph{Forward Process and Feature Projection.}
We adopt the standard Gaussian diffusion formulation~\cite{DBLP:conf/nips/HoJA20}. The forward process $q(z_n|z_0)$ imposes noise according to a variance schedule $\{\beta_n\}_{n=1}^N$:
\begin{equation}
    z_n = \sqrt{\bar{\alpha}_n}z_0 + \sqrt{1-\bar{\alpha}_n}\epsilon, \quad \epsilon \sim \mathcal{N}(0, \mathbf{I}),
\end{equation}
where $\alpha_n = 1 - \beta_n$ and $\bar{\alpha}_n = \prod_{s=1}^n \alpha_s$. To leverage the inductive biases of vision transformers, we project the unstructured vector $z_n$ into a pseudo-spatial feature map. We define a learnable linear projector $\mathcal{P}: \mathbb{R}^{d_{\mathrm{VL}}} \to \mathbb{R}^{C \times H \times W}$ with $(C, H, W) = (4, 32, 32)$, such that:
\begin{equation}
    \mathbf{H}_n = \mathcal{P}(z_n) \in \mathbb{R}^{4 \times 32 \times 32}.
\end{equation}
This mapping $\mathcal{P}$ transforms the fused VL embedding into a structured tensor $\mathbf{H}_n$, analogous to a latent feature map, enabling patch-based processing.

\paragraph{Transformer Backbone Formulation.}
The backbone processes $\mathbf{H}_n$ via patchification. We partition $\mathbf{H}_n$ into non-overlapping patches of size $p \times p$ (where $p=2$), resulting in a sequence of $M = (H/p)(W/p) = 256$ tokens. These are linearly projected to a hidden dimension $D=1152$ and augmented with sinusoidal positional embeddings $\mathbf{E}_{\mathrm{pos}}$:
\begin{equation}
    \mathbf{X}_0 = \mathrm{Linear}(\mathrm{Patchify}(\mathbf{H}_n)) + \mathbf{E}_{\mathrm{pos}} \in \mathbb{R}^{M \times D}.
\end{equation}
The sequence $\mathbf{X}_0$ propagates through a stack of $L=28$ transformer blocks. Each block $\ell$ incorporates timestep information $n$ and textual conditions $c_T$.

\paragraph{Conditioning Mechanisms.}
\textbf{Time Modulation:} The timestep $n$ is mapped to an embedding $\tau \in \mathbb{R}^D$. We employ Adaptive Layer Normalization (AdaLN), where $\tau$ regresses scale ($\gamma$) and shift ($\beta$) parameters to modulate normalized features:
\begin{equation}
    \mathrm{AdaLN}(\mathbf{X}, n) = \gamma(\tau) \odot \mathrm{LN}(\mathbf{X}) + \beta(\tau).
\end{equation}
\textbf{Text Conditioning:} The caption is encoded into a sequence $\mathbf{C}_{\mathrm{text}} \in \mathbb{R}^{K \times D}$ via the text branch of $f_M$ and a projection MLP. Interaction occurs via Multi-Head Cross-Attention (MHCA).

\paragraph{Block Architecture.}
The operation of the $\ell$-th transformer block is defined by the composition of Multi-Head Self-Attention (MHSA), MHCA, and a Feed-Forward Network (FFN):
\begin{align}
    \mathbf{X}'_\ell &= \mathbf{X}_{\ell-1} + \mathrm{MHSA}\bigl(\mathrm{AdaLN}(\mathbf{X}_{\ell-1}, n)\bigr), \\
    \mathbf{X}''_\ell &= \mathbf{X}'_\ell + \mathrm{MHCA}\bigl(\mathrm{AdaLN}(\mathbf{X}'_\ell, n), c_T\bigr), \\
    \mathbf{X}_\ell &= \mathbf{X}''_\ell + \mathrm{FFN}\bigl(\mathrm{AdaLN}(\mathbf{X}''_\ell, n)\bigr).
\end{align}
To balance efficiency and receptive field, we alternate MHSA between global attention and local windowed attention (window size $16\times16$). Zero-initialized gating is applied to the residual branches of each sub-layer for stable training dynamics.

\paragraph{Output Parameterization.}
The final hidden states $\mathbf{X}_L$ are reshaped via a linear unpatchify operation $\mathcal{P}^{-1}$ to recover the VL dimensionality. The network predicts the noise component $\epsilon_\theta$:
\begin{equation}
    \epsilon_\theta(z_n, n, c_T) = \mathrm{Linear}_{\mathrm{out}}\bigl(\mathcal{P}^{-1}(\mathbf{X}_L)\bigr) \in \mathbb{R}^{d_{\mathrm{VL}}}.
\end{equation}
The model is optimized to minimize the simple diffusion objective $\|\epsilon - \epsilon_\theta(z_n, n, c_T)\|^2$ directly in the joint embedding space.

\subsection{Training Strategy and Hyperparameters}

\paragraph{Diffusion Prior Pre-training.}
We instantiate the diffusion backbone using the DiT-XL/2 architecture configuration. To adapt the model for global vision--language feature modeling, we disable local windowed attention and relative positional encodings. Furthermore, we enforce full-precision attention computation to ensure numerical precision during the diffusion process.
The model is optimized using AdamW~\cite{DBLP:journals/corr/KingmaB14} with a global batch size of $4096$. We employ the following hyperparameter settings:
\begin{itemize}
    \item Learning rate: $\eta = 2 \times 10^{-5}$;
    \item Weight decay: $\lambda = 3 \times 10^{-2}$;
    \item Optimizer stability term: $\varepsilon = 10^{-10}$.
\end{itemize}
To stabilize training dynamics under this large effective batch size, we apply global gradient clipping with an $\ell_2$-norm threshold of $0.01$.
The model is trained for a total of $4$ epochs on the Stage 1 corpus. We utilize a learning rate warmup strategy for the initial $1000$ optimization steps, followed by a cosine annealing schedule for the remainder of the training process.

\paragraph{Feature Editing Fine-tuning.}
In the second stage, we freeze the parameters of the pre-trained diffusion backbone to preserve the robust generative prior learned from large-scale pre-training. We instantiate the Control-Adapter by duplicating the weights of the DiT encoder blocks. To ensure a stable optimization start, the zero-convolution layers (linear projections connecting the adapter to the backbone) are initialized with zeros, ensuring the initial forward pass remains identical to the pre-trained prior. The model is fine-tuned on the synthetic HQEdit dataset containing approximately $200\mathrm{K}$ triplets.
We continue to use the AdamW optimizer~\cite{DBLP:journals/corr/KingmaB14} for fine-tuning. Given the introduction of the Control-Adapter and the reduced dataset size, we adopt the following specific settings:
\begin{itemize}
    \item Learning rate: $\eta = 2 \times 10^{-5}$ (typically higher for adapter training);
    \item Weight decay: $\lambda = 2 \times 10^{-2}$;
\end{itemize}
Training is distributed across $8$ NVIDIA A100 GPUs. To prevent overfitting to the synthetic edit patterns, we maintain the global gradient clipping threshold at $0.01$.
The Control-Adapter is trained for $2$ epochs on the HQEdit dataset. We employ a constant learning rate schedule without warmup, as the Control-Adapter is initialized from pre-trained weights and the zero-initialization strategy mitigates the need for a gradual warmup phase.

\section{Training Cost and Inference Speed}

We provide a detailed analysis of the computational resources, training costs, and inference latency associated with Fusion-Diff. Table~\ref{tab:training-cost} summarizes the configuration and efficiency metrics. It is important to note that Stage 1 is a pure pre-training phase used solely to align the diffusion prior with the joint vision--language space. The actual retrieval inference is performed only after the feature editing phase, utilizing the fine-tuned Control-Adapter.

\paragraph{Training Efficiency.}
All models were trained on a node equipped with $8 \times$ NVIDIA A100 (80GB) GPUs.
Pre-training is the most computationally intensive phase, requiring approximately 80 GPU-hours to align the DiT-XL/2 backbone (Total Depth $2L=28$) using the 33M sample corpus.
In contrast, Stage 2 is highly efficient; by freezing the backbone and training only the lightweight Control-Adapter on 200K synthetic triplets, we reduce the training time to approximately 0.43 GPU-hours. This demonstrates that once the generic prior is established, adapting Fusion-Diff to the composed image retrieval task is computationally inexpensive.

\paragraph{Accelerated Inference via DPM-Solver.}
During the inference phase (Stage 2), standard diffusion sampling (e.g., DDPM~\cite{DBLP:conf/nips/HoJA20}) typically necessitates a large number of iterative denoising steps, introducing significant latency. To enable practical deployment, we employ DPM-Solver++~\cite{DBLP:journals/ijautcomp/LuZBCLZ25}, a fast high-order solver for diffusion ODEs. Specifically, we utilize a second-order solver with a reduced number of inference steps ($N=14$). This optimization drastically reduces the per-query inference latency to approximately 0.32 seconds (excluding MLLM encoding time) with negligible degradation in retrieval accuracy.

\begin{table*}[t]
\centering
\caption{Hyperparameters, training details, and compute resources for Fusion-Diff. We report metrics separately for the generic Pre-training and the task-specific Feature Editing. Note that Stage 1 is training-only; inference latency is reported only for the final model, where we utilize the DPM-Solver++ to accelerate retrieval.}
\label{tab:training-cost}
\begin{tabular}{lcc}
\toprule
Parameter / Metric & \textbf{Stage 1: Pre-training} & \textbf{Stage 2: Feature Editing} \\
\midrule
\textbf{Architecture (DiT-XL/2)} & & \\
Hidden dimension ($d$) & 1152 & 1152 \\
Attention heads & 16 & 16 \\
Encoder/Decoder Blocks ($L$) & 14 & 14 \\
Patch size ($p$) & 2 & 2 \\
Trainable Parameters & 954M & 318M \\
\midrule
\textbf{Training Configuration} & & \\
Global Batch Size & 4096 & 1024 \\
Optimizer & AdamW & AdamW \\
Learning Rate & $2 \times 10^{-5}$ & $2 \times 10^{-5}$ \\
Data Scale & 33M (Image-Text Pairs) & 200K (Synthetic Triplets) \\
Total Training Steps & $\sim$32800 & $\sim$400 \\
Approx. Training Time & $\sim$80 hours (4 epochs) & $\sim$26 mins (2 epochs) \\
\midrule
\textbf{Inference \& Efficiency} & & \\
Status & \textit{N/A (Training Only)} & Inference \\
Sampler Strategy & -- & DPM-Solver++ \\
Inference Steps ($N$) & -- & 14 \\
Guidance Scale ($\gamma$) & -- & 2.5 \\
Control Scale ($\delta$) & -- & 1.0 \\
Inference Latency (Diffusion) & -- & $\sim$0.32 s / query \\
\bottomrule
\end{tabular}
\end{table*}

\section{Algorithm}
In this section, we present the comprehensive algorithmic framework of Fusion-Diff, encompassing four key components: (1) VL-Embed (Algorithm~\ref{alg:vl-embed}): The construction of the joint vision--language manifold using the MLLM; (2) the Stage 1 Pre-training procedure (Algorithm~\ref{alg:stage1-train}), where we learn a conditional diffusion prior over fused multimodal embeddings; (3) the Stage 2 Fine-tuning procedure (Algorithm~\ref{alg:stage2-train}), which freezes the prior and trains the Control-Adapter to perform query-specific editing; and (4) Fusion-Retrieval (Algorithm~\ref{alg:inference}): the complete zero-shot inference pipeline that generates target features and ranks gallery images.

\begin{algorithm}[htbp]
\caption{VL-Embed: Joint Vision-Language Space Construction}
\label{alg:vl-embed}
\begin{algorithmic}[1]
\Require Image-Text Pair Dataset $\mathcal{D}_{\text{pair}}$, MLLM $f_M$
\Ensure Latent Dataset $\mathcal{D}_{\text{latent}} = \{(z_0, c_T)\}$
\State $\mathcal{D}_{\text{latent}} \gets \emptyset$
\For{each pair $(I, T) \in \mathcal{D}_{\text{pair}}$}
    \State \textbf{// Step 1: Construct Multimodal Input}
    \State $x_{\text{pair}} \gets \text{FormatInput}(I, T)$
    \State \textbf{// Step 2: Extract Fused Embedding}
    \State $z_0 \gets f_M(x_{\text{pair}})$ \Comment{Target fused feature in $\mathbb{R}^{d_{\text{VL}}}$}
    \State $z_0 \gets \text{Normalize}(z_0)$
    \State \textbf{// Step 3: Extract Text Condition}
    \State $c_T \gets \mathcal{E}_{\text{text}}(T)$ \Comment{Caption embedding}
    \State $\mathcal{D}_{\text{latent}} \gets \mathcal{D}_{\text{latent}} \cup \{(z_0, c_T)\}$
\EndFor
\State \Return $\mathcal{D}_{\text{latent}}$
\end{algorithmic}
\end{algorithm}

\begin{algorithm}[htbp]
\caption{Stage 1: Diffusion Prior Pre-training}
\label{alg:stage1-train}
\begin{algorithmic}[1]
\Require Latent Dataset $\mathcal{D}_{\text{latent}}$, Diffusion Model $\epsilon_\theta$
\Ensure Pre-trained Parameters $\Theta_{\text{prior}}$
\State Initialize $\Theta_{\text{prior}}$ randomly
\For{epoch $= 1$ to $N_{\text{epochs}}$}
    \For{$(z_0, c_T) \in \mathcal{D}_{\text{latent}}$}
        \State $n \sim \text{Uniform}(1, N)$ \Comment{Sample timestep $n$}
        \State $\epsilon \sim \mathcal{N}(0, \mathbf{I})$
        \State $z_n \gets \sqrt{\bar{\alpha}_n}z_0 + \sqrt{1-\bar{\alpha}_n}\epsilon$ \Comment{Eq. 2}
        \State \textbf{// Classifier-Free Guidance Training}
        \State $u \sim \text{Uniform}(0, 1)$
        \If{$u < p_{\text{cfg}}$}
            \State $c_{\text{cond}} \gets \emptyset$ 
        \Else
            \State $c_{\text{cond}} \gets c_T$
        \EndIf
        \State $\epsilon_{\text{pred}} \gets \epsilon_\theta(z_n, n, c_{\text{cond}})$
        \State $\mathcal{L}_{\text{Stage1}} \gets \|\epsilon - \epsilon_{\text{pred}}\|_2^2$ \Comment{Eq. 3}
        \State Update $\Theta_{\text{prior}}$ using $\nabla \mathcal{L}_{\text{Stage1}}$
    \EndFor
\EndFor
\end{algorithmic}
\end{algorithm}

\begin{algorithm}[htbp]
\caption{Stage 2: Feature Editing with Control-Adapter}
\label{alg:stage2-train}
\begin{algorithmic}[1]
\Require Triplet Dataset $\mathcal{D}_{\text{edit}}$, Pre-trained Prior $\epsilon_\theta$
\Ensure Trained Control-Adapter Parameters $\Theta_{\text{ctrl}}$
\State \textbf{// Initialize Control-Adapter}
\State Freeze $\Theta_{\text{prior}}$. Initialize $\Theta_{\text{ctrl}}$ from copy/zeros.
\For{iteration $= 1$ to $N_{\text{iter}}$}
    \State Sample $(I_r, T_\Delta, I_t, T_t) \sim \mathcal{D}_{\text{edit}}$
    \State $z_{r,\Delta} \gets f_M(I_r, T_\Delta)$ \Comment{Query condition}
    \State $z_t \gets f_M(I_t, T_t)$ \Comment{Target Ground Truth}
    \State $c_\Delta \gets \mathcal{E}_{\text{text}}(T_\Delta)$ \Comment{Text condition}
    \State $n \sim \text{Uniform}(1, N), \epsilon \sim \mathcal{N}(0, \mathbf{I})$
    \State $z_{n,\text{noise}} \gets \text{AddNoise}(z_t, n, \epsilon)$
    \State \textbf{// Control-Adapter Forward}
    \State Compute control features $\{y_c\}$ using $z_{r,\Delta}$ and $z_{n,\text{noise}}$
    \State $\epsilon_{\text{pred}} \gets \epsilon_\theta(z_{n,\text{noise}}, n, c_\Delta, \{y_c\})$
    \State $\mathcal{L}_{\text{Stage2}} \gets \|\epsilon - \epsilon_{\text{pred}}\|_2^2$ \Comment{Eq. 5}
    \State Update $\Theta_{\text{ctrl}}$ using $\nabla \mathcal{L}_{\text{Stage2}}$
\EndFor
\end{algorithmic}
\end{algorithm}

\begin{algorithm}[htbp]
\caption{Fusion-Retrieval: Inference Pipeline}
\label{alg:inference}
\begin{algorithmic}[1]
\Require Query $(I_r, T_\Delta)$, Gallery $\mathcal{D}_{\text{gal}}$
\Ensure Ranked List of Images
\State $z_{r,\Delta} \gets f_M(I_r, T_\Delta)$
\State $\tilde{T} \gets \text{GenerateDescription}(f_M, I_r, T_\Delta)$
\State $c_{\tilde{T}} \gets \mathcal{E}_{\text{text}}(\tilde{T})$
\State $\hat{z}_N \sim \mathcal{N}(0, \mathbf{I})$ \Comment{Start from noise at $n=N$}
\For{$n = N$ down to $1$ (or DPM steps)}
    \State $\epsilon_{\text{cond}} \gets \epsilon_\theta(\hat{z}_n, n, c_{\tilde{T}}, z_{r,\Delta})$
    \State $\epsilon_{\text{uncond}} \gets \epsilon_\theta(\hat{z}_n, n, \emptyset, \emptyset)$
    \State $\tilde{\epsilon} \gets (1+\gamma)\epsilon_{\text{cond}} - \gamma\epsilon_{\text{uncond}}$ \Comment{Eq. 4}
    \State $\hat{z}_{n-1} \gets \text{SolverStep}(\hat{z}_n, \tilde{\epsilon}, n)$
\EndFor
\State \textbf{// Retrieval}
\State Calculate Cosine Similarity between $\hat{z}_0$ and all $z_g \in \mathcal{D}_{\text{gal}}$
\State \Return Top-ranked images
\end{algorithmic}
\end{algorithm}

\section{More Qualitative Results}

To provide a more comprehensive assessment of Fusion-Diff, we present additional qualitative comparisons against two state-of-the-art zero-shot baselines: CIReVL~\cite{DBLP:conf/iclr/KarthikRMA24} (a representative training-free, LLM-reasoning method) and LinCIR+CIG~\cite{DBLP:conf/cvpr/WangABL25} (a representative generative method that synthesizes pseudo-images).
We visualize the retrieval results (Recall@1) on the CIRR and FashionIQ benchmarks. These examples further substantiate our claim that modeling the conditional distribution in the joint vision--language space yields superior alignment compared to text-centric reasoning or pixel-space generation.

\subsection{More Qualitative results on CIRR}

\begin{figure*}[h]
    \centering
    \includegraphics[width=1.0\linewidth]{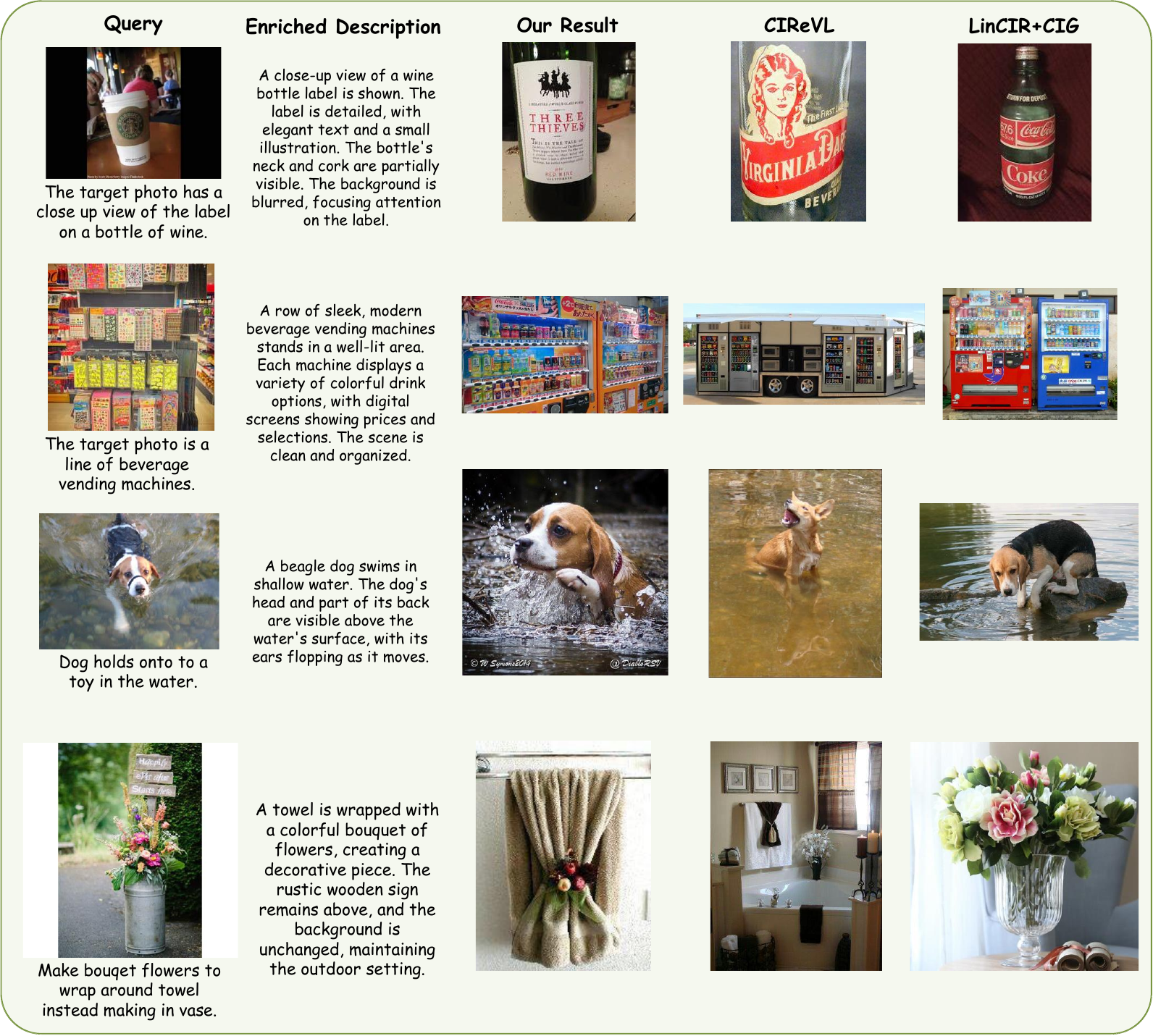}
    \caption{Additional qualitative comparison on the \textbf{CIRR} validation set. We compare the Recall@1 retrieved images of our Fusion-Diff against CIReVL~\cite{DBLP:conf/iclr/KarthikRMA24} and LinCIR+CIG~\cite{DBLP:conf/cvpr/WangABL25}. Fusion-Diff accurately captures fine-grained visual details and spatial relationships where baselines suffer from semantic drift or hallucination.}
    \label{fig:supp_cirr}
\end{figure*}

Figure~\ref{fig:supp_cirr} illustrates qualitative results on the CIRR validation set. The queries in CIRR often involve complex spatial modifications or state changes that are difficult to capture via text alone.

\textbf{Analysis of Semantic Consistency.}
In the first row (Wine Bottle), the query demands a specific visual viewpoint change (``close up view of the label'').
LinCIR+CIG suffers from severe hallucination, retrieving a carbonated soft drink bottle which is visually distinct from the requested wine bottle category. This failure likely stems from the noise inherent in pixel-space diffusion when the textual condition (``bottle'') is broad.
CIReVL, relying heavily on textual reasoning, retrieves a bottle with a completely different label structure and color palette compared to the ground truth.
In contrast, Fusion-Diff successfully retrieves the correct target, demonstrating that our joint-space diffusion preserves the fine-grained visual identity and layout required by the query while faithfully executing the viewpoint change.

\textbf{Analysis of Complex Composition.}
The fourth row (Flowers and Towel) presents a challenging compositional query: ``wrap around towel instead making in vase".
CIReVL fails to capture the ``wrapping" action, retrieving a standard bathroom interior.
LinCIR+CIG retrieves a generic vase, ignoring the modification instruction entirely.
Fusion-Diff is the only method that retrieves the correct image where the bouquet is tied around the towel. This highlights the effectiveness of the \textbf{Control-Adapter}, which injects the structural layout of the query directly into the diffusion process, enabling precise compositional reasoning that pure language models miss.

\subsection{More Qualitative results on FashionIQ}

\begin{figure*}[h]
    \centering
    \includegraphics[width=1.0\linewidth]{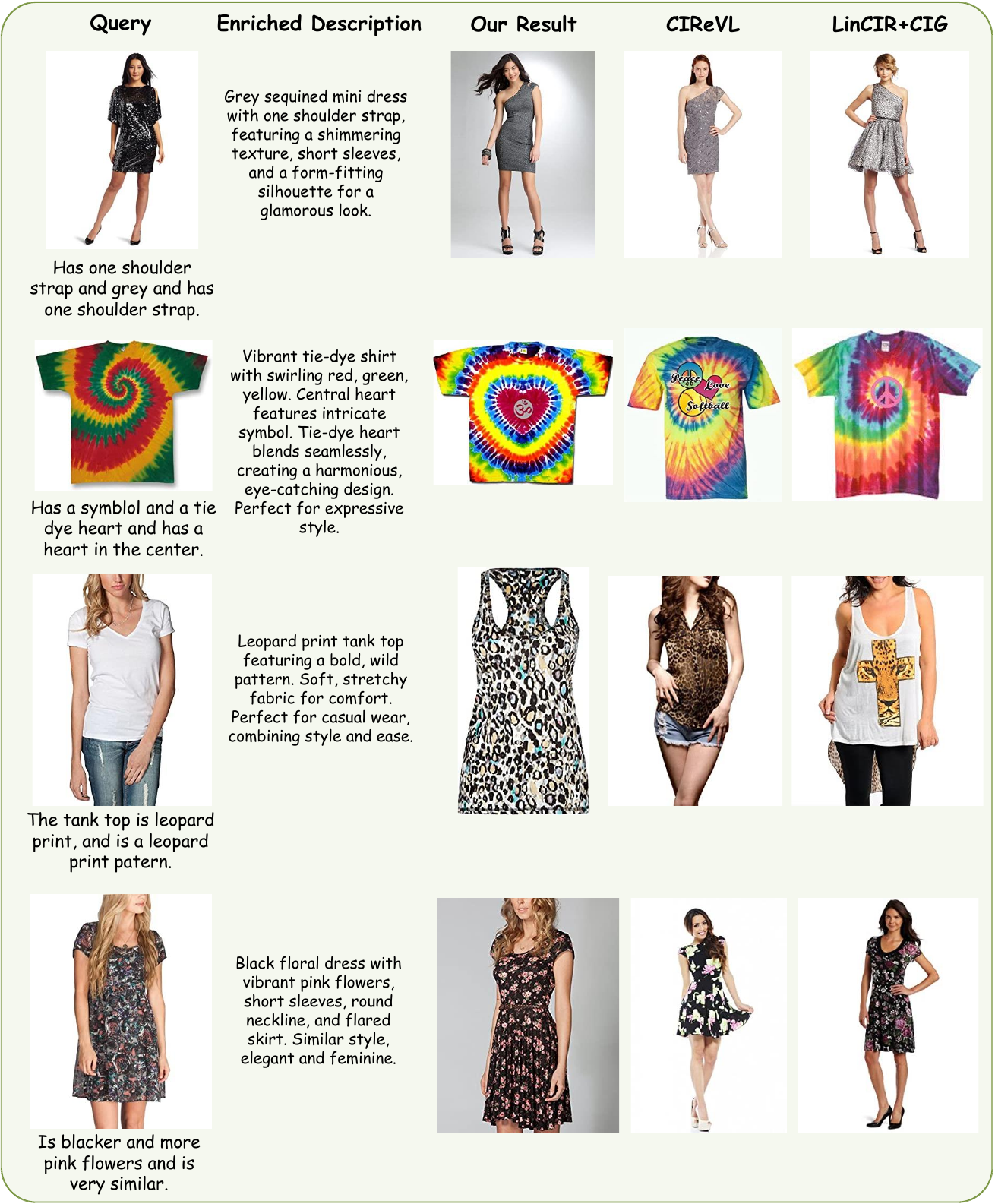}
    \caption{Additional qualitative comparison on the \textbf{FashionIQ} validation set. Fusion-Diff demonstrates superior capability in following fine-grained attribute editing instructions (e.g., specific logos, patterns, and cuts) compared to CIReVL~\cite{DBLP:conf/iclr/KarthikRMA24} and LinCIR+CIG~\cite{DBLP:conf/cvpr/WangABL25}.}
    \label{fig:supp_fiq}
\end{figure*}

Figure~\ref{fig:supp_fiq} showcases results on the FashionIQ validation set, focusing on fine-grained attribute manipulation (e.g., texture, pattern, and color changes).

\textbf{Pattern and Symbol Alignment.}
The second row (Tie-dye Shirt) provides a compelling example of fine-grained pattern recognition. The edit instruction explicitly asks for a ``heart in the center". LinCIR+CIG retrieves a shirt with a peace sign, illustrating a limitation where the synthesized pseudo-visual features introduce noise that misdirects the retrieval process, causing the model to capture the general style (tie-dye) but fail to align with the specific symbol requirement.
CIReVL retrieves a shirt with text (``Peace Love Softball"), failing to align with the visual concept of a ``heart".
Fusion-Diff accurately retrieves the shirt featuring a heart pattern. This precision confirms that our method, by operating in the joint VL space, avoids the modality gap where text descriptions fail to fully encapsulate complex visual patterns.

\textbf{Attribute and Silhouette Fidelity.}
In the first row (Grey Dress), the query specifies ``one shoulder strap".
LinCIR+CIG retrieves a dress with a completely different silhouette (flared skirt, standard straps).
CIReVL retrieves a dress that matches the ``one shoulder" attribute but misses the specific texture (sequins) and fit.
Fusion-Diff retrieves the exact match, preserving the tight-fitting silhouette and the metallic texture of the reference while satisfying the structural edit.

Across both datasets, comparisons reveal that Fusion-Diff significantly reduces two types of errors prevalent in baselines: (1) the hallucination of irrelevant objects seen in generative methods (LinCIR+CIG), and (2) the semantic drift seen in text-centric methods (CIReVL) where visual constraints are lost. These qualitative results strongly support the quantitative gains reported in the main paper.

\clearpage
\newpage
{
    \small
    \bibliographystyle{ieeenat_fullname}
    \bibliography{main}
}


\end{document}